%% file: main.tex
\documentclass[10pt,twocolumn,letterpaper]{article}

\usepackage[final]{cvpr}      

\usepackage{graphicx}
\usepackage{amsmath}
\usepackage{amssymb}
\usepackage{booktabs}

\usepackage[pagebackref,breaklinks,colorlinks]{hyperref}

\usepackage[capitalize]{cleveref}
\crefname{section}{Sec.}{Secs.}
\Crefname{section}{Section}{Sections}
\Crefname{table}{Table}{Tables}
\crefname{table}{Tab.}{Tabs.}

\def\cvprPaperID{*****} 
\def\confName{CVPR}
\def\confYear{2022}

\begin{document}

\title{All Robots in One: A New Standard and Unified Dataset for Versatile, General-Purpose Embodied Agents}

\author{
{\fontsize{11.0pt}{\baselineskip}\selectfont
Zhiqiang Wang${}^{2}$\footnotemark[1],
Hao Zheng${}^{2}$\footnotemark[1],
Yunshuang Nie${}^{3}$\footnotemark[1],
Wenjun Xu${}^{1}$\footnotemark[1],
Qingwei Wang${}^{2}$\footnotemark[1],
Hua Ye${}^{1}$\footnotemark[1],
Zhe Li${}^{2}$},\\
{\fontsize{11.0pt}{\baselineskip}\selectfont
Kaidong Zhang${}^{3}$,
Xuewen Cheng${}^{1}$, 
Wanxi Dong${}^{2}$,
Chang Cai${}^{1}$,
Liang Lin${}^{1,3}$,
Feng Zheng${}^{1,2}$\footnotemark[2],
Xiaodan Liang${}^{1,3}$\footnotemark[2]}\\
{\fontsize{11.0pt}{\baselineskip}\selectfont
${}^1$Pengcheng Laboratory 
${}^2$Southern University of Science and Technology 
${}^3$Sun Yat-sen University}\\
}

\maketitle

\input{S0_Abstract}

\renewcommand{\thefootnote}{\fnsymbol{footnote}}
\footnotetext[1]{Equal Contribution}
\footnotetext[2]{Corresponding Author}
\input{S1_Introduction}

\input{S2_RelatedWork}
\input{S3_ARIO}

\input{S4_Discussion}

\input{S5_Acknowledgement}

{\small
\bibliographystyle{ieee_fullname}
\bibliography{egbib}
}

\end{document}

%% file: S0_Abstract.tex
\begin{abstract}
   Embodied AI is transforming how AI systems interact with the physical world, yet existing datasets are inadequate for developing versatile, general-purpose agents.
   These limitations include a lack of standardized formats, insufficient data diversity, and inadequate data volume.
   To address these issues, we introduce ARIO (All Robots In One), a new data standard that enhances existing datasets by offering a unified data format, comprehensive sensory modalities, and a combination of real-world and simulated data. 
   ARIO aims to improve the training of embodied AI agents, increasing their robustness and adaptability across various tasks and environments.
   Building upon the proposed new standard, we present a large-scale unified ARIO dataset, comprising approximately 3 million episodes collected from 258 series and 321,064 tasks.
   The ARIO standard and dataset represent a significant step towards bridging the gaps of existing data resources. 
   By providing a cohesive framework for data collection and representation, ARIO paves the way for the development of more powerful and versatile embodied AI agents, capable of navigating and interacting with the physical world in increasingly complex and diverse ways. 
   The project is available on \url{https://imaei.github.io/project_pages/ario/}.
\end{abstract}

%% file: S1_Introduction.tex
\input{Figures/cover}

\section{Introduction}
Embodied AI is now significantly impacting the way AI systems interact with the physical world by integrating perception, cognition, and action. This development is driving advances across diverse fields, from robotics to human-computer interaction, highlighting the need for comprehensive and versatile datasets. Several prior studies have introduced open-source datasets designed for specific tasks such as grasping, routing, and pick-and-place, aiming to train agents tailored for particular scenarios. Open X-embodiment \cite{open_x_embodiment_rt_x_2023} further aggregates data from various datasets, spanning multiple robotic platforms, tasks and environments, to facilitate large-scale robotic pre-training.

However, substantial limitations at the data level continue to hinder the development of robust, general-purpose embodied agents, particularly in terms of standardized formats, diversity, and data volume. Task-specific datasets are inadequate for training these versatile agents, and despite the seemingly unified structure of pre-training datasets like Open X-embodiment, significant issues still persist. These issues include a lack of comprehensive sensory modalities, with no dataset incorporating images, 3D vision, text, tactile, and auditory inputs simultaneously; the absence of a unified format in multi-robot datasets, complicating data processing and loading; incompatibility in representing diverse control objects across different robotic platforms; insufficient data volume which impedes large-scale pretraining; and a shortage of datasets that combine simulated and real data, which is crucial for studying the sim-to-real gap.

To address these challenges, we first introduce ARIO (All Robots In One), a new dataset standard that comprehensively optimizes existing datasets, enabling the development of more versatile and general-purpose embodied AI agents.
In the ARIO standard, control and motion data from robots of different morphologies are recorded in a unified format. A timestamp mechanism is incorporated to standardize data collection, addressing the variability in robot action frequencies and sensor frame rates (See in Section 3.1).
ARIO's unified format allows for variable data from different robot types, ensuring accurate timestamps. This enables users to efficiently train high-performing, generalizable embodied AI models, making ARIO standard an ideal format for embodied AI datasets.

Building upon the ARIO standard, we have constructed a unified large-scale ARIO dataset, encompassing approximately 3 million episodes collected from 258 series and 321,064 tasks. This dataset was assembled through a multi-pronged approach: (1) real-world data collection from our custom platform, yielding 3,662 episodes across over 105 tasks, (2) simulation-based data generation using various platforms (Habitat, MuJoCo, and SeaWave), resulting in 703,088 episodes from 1198 tasks, and (3) conversion of existing open-source datasets to the ARIO standard, contributing 2,326,438 episodes from 319,761 tasks. The ARIO dataset overcomes the limitations of current datasets and paves the way for the development of robust, general-purpose embodied agents.

The key advantages of the ARIO standard and dataset can be summarized as follows:
\begin{itemize}
    \setlength\itemsep{0.1em}
    \item \textbf{Comprehensive Sensory Modalities}: ARIO incorporates five sensory modalities—images, 3D data, sound, text, and tactile information—offering a richer and more diverse dataset.
    \item \textbf{Timestamp-Based Data Alignment}: Data is recorded and synchronized based on timestamps, accommodating different sensor frame rates.
    \item \textbf{Structured Series-Task-Episode Framework}: ARIO uses a clear series-task-episode structure, with detailed textual descriptions for each series and task.
    \item \textbf{Unified Data Format}: The dataset follows a standardized format, supporting various robot types and control objects, simplifying data processing.
    \item \textbf{Integration of Simulation and Real Robot Data}: ARIO includes both simulation and real-world data across multiple tasks, scenarios, and robot types, enhancing generalization across different hardware platforms.
    \item \textbf{Cleaning and Standardization}: ARIO cleans and standardizes large open-source datasets, making them easier to use and integrate.
\end{itemize}

%% file: Figures/cover.tex
\begin{figure*}[t]
  \centering
   \includegraphics[width=0.95\linewidth]{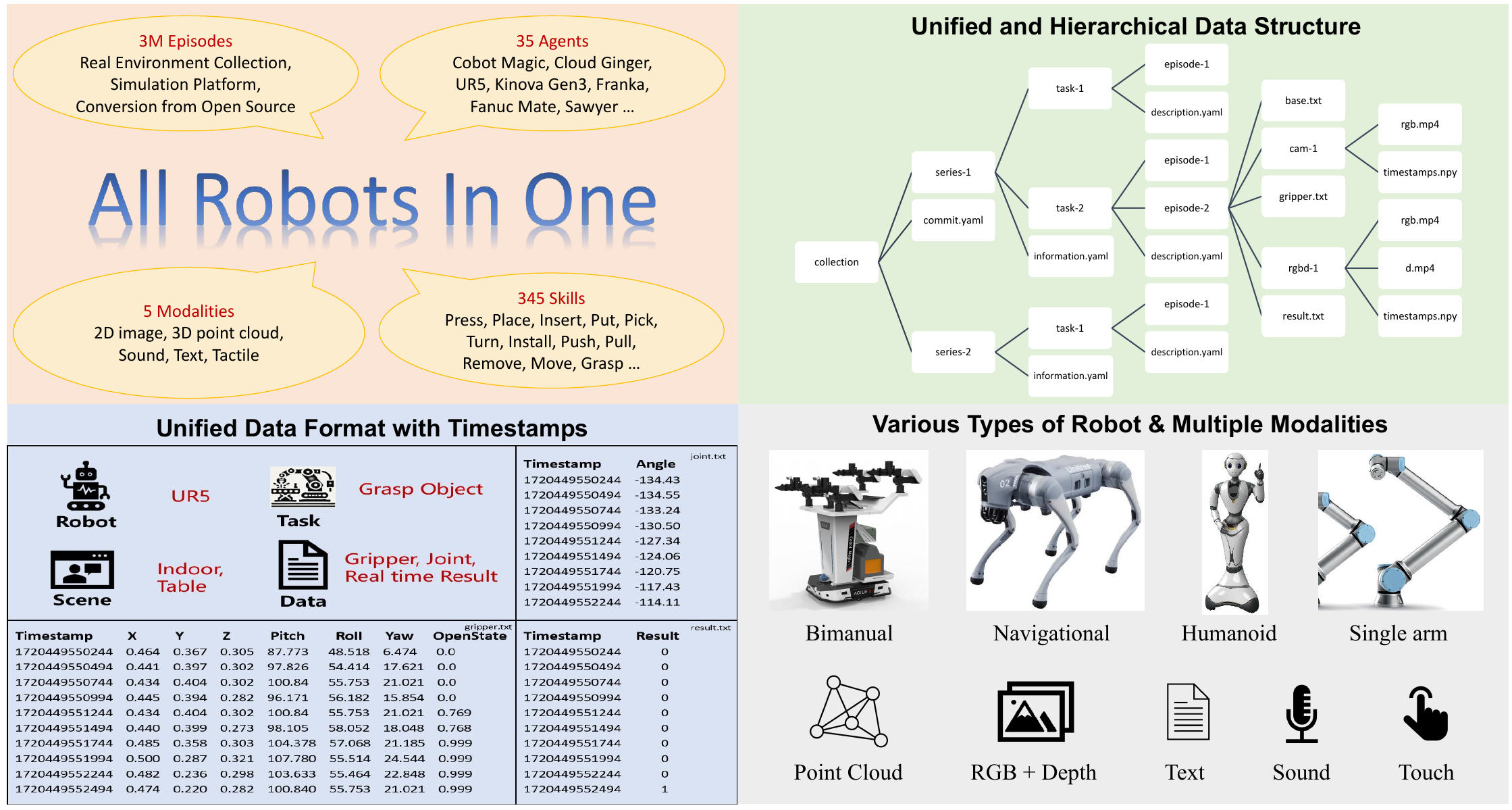}
   \caption{All robots in one.}
   \label{fig:cover}
\end{figure*}

%% file: S2_RelatedWork.tex
\section{Related Work}

\textbf{Large-Scale Robotic Learning Datasets:}
In response to the growing need for large-scale, diverse datasets within the robotics community, numerous datasets have been developed, primarily focusing on robotic manipulation tasks such as grasping, pushing, and object interactions~\cite{brohan2022rt, dasari2019robonet, forbes2014robot, jang2021bc, kalashnikov2021mt, mandlekar2018roboturk, sharma2018multiple}. Notably, datasets like RoboNet~\cite{dasari2019robonet}, RT-1~\cite{brohan2022rt}, and BC-Z~\cite{jang2021bc} have made significant contributions by aggregating manipulation data across varied robotic setups and environments. However, these datasets are often limited in task diversity and the richness of sensory data, with a primary emphasis on visual data, while largely neglecting modalities such as tactile, auditory, or proprioceptive feedback. The ARIO dataset addresses these limitations by incorporating five sensory modalities—image, 3D vision, audio, text, and tactile—thus enriching the dataset for training robust multi-modal perception models.

\textbf{Multi-Embodiment and Cross-Platform Learning:}
Transferring knowledge across diverse robotic platforms remains a formidable challenge. Prior research~\cite{chen2018hardware, huang2020one, ghadirzadeh2021bayesian, xu2021adagrasp, shao2020unigrasp, zhou2023learning, hu2021know} has explored the use of datasets from various robotic platforms to enhance generalization and facilitate positive transfer. These studies have sought to bridge the embodiment gap through shared action representations and the separation of robot and environment characteristics. However, existing datasets, such as Open X-Embodiment~\cite{open_x_embodiment_rt_x_2023}, often lack standardized formats, which complicates data processing and utilization across platforms. The ARIO dataset addresses these challenges by standardizing data formats and including both simulated and real-world data, offering a unique framework for examining the sim-to-real transition, which is crucial for developing versatile robotic systems.

\textbf{Multi-Modal Learning and Sensory Integration:}
The integration of diverse sensory modalities is crucial for executing complex, contact-rich manipulation tasks. While traditional datasets have primarily focused on visual inputs, recent initiatives have started to incorporate additional sensory data, such as tactile feedback~\cite{lee2019making, liu2024maniwav, kerr2022self, chen2022visuo, wi2022virdo++, guzey2023dexterity, liu2023enhancing}. For example, the ManiWAV dataset~\cite{liu2024maniwav} introduced audio feedback as a novel modality to detect contact events, showcasing the potential of acoustic sensing in robotic learning. By integrating multiple sensory modalities, ARIO provides a richer dataset that enables more sophisticated and robust robotic perception and control.

\textbf{Language-Guided Robot Learning:}
Language-guided learning is emerging as a powerful approach to scale robotic skill acquisition. This method leverages large language models (LLMs) to guide both data collection and task execution~\cite{shridhar2023perceiver, shridhar2022cliport, lynch2020language, stepputtis2020language, jang2021bc, lynch2023interactive, mees2022matters, brohan2022rt}. Human annotations play a crucial role in this process, providing both action labels and natural language descriptions. Datasets used in this domain range from task-specific data~\cite{shridhar2022cliport, shridhar2023perceiver, jang2021bc, brohan2022rt} to task-agnostic "play" data~\cite{lynch2020language, stepputtis2020language, lynch2023interactive, mees2022calvin}. However, scalability remains a challenge, as the data collection process largely depends on human input. ARIO contributes to this field by providing a standardized, multi-modal dataset that can support language-guided learning, particularly in diverse and complex embodied tasks.

%% file: S3_ARIO.tex
\section{ARIO: All Robots In One}
\subsection{What is the ARIO Standard?}

The ARIO standard is a framework that standardizes the collection, storage, and analysis of embodied AI data across various environments and tasks. Its modular design supports scalable robot foundation model design and effective algorithm testing.

\textbf{Hierarchical Data Structure:} ARIO organizes data into four principal layers: collection, series, task, and episode. A collection encompasses multiple series, each aligned with a specific scene and robot type, while a series consists of multiple tasks described by natural language instructions like "picking an apple." Tasks are subdivided into episodes, each capturing a complete set of data from a single execution, including all observation and control data synchronized by unified timestamps.

\textbf{Data Collection Protocols:} To capture a broad operational spectrum, ARIO mandates collecting a wide array of environments and actions. Each data capture session records essential modalities like text instructions and images, along with task-specific modalities such as end-effector states or navigational data.

\textbf{Metadata and Documentation:} Comprehensive metadata is provided through \texttt{information.yaml} files in each series, detailing the involved scene, robot, and sensors, ensuring alignment with the data in respective episodes. Task-specific metadata is contained in \texttt{description.yaml} files, outlining detailed instructions and required skills for each task.

\textbf{Standardization and Integrity:} ARIO emphasizes standardized data formats and meticulous collection protocols across various sensors and interactions to ensure data integrity and usability. This standardization supports straightforward data integration and analysis, crucial for developing adaptable and scalable robotic agents.

By providing a structured data handling approach, the ARIO standard significantly enhances the capacity for robotic learning and generalization across diverse scenarios and tasks.

\begin{figure}[htbp]
    \centering
    \begin{subfigure}[b]{0.23\textwidth}
        \centering
        \includegraphics[width=\textwidth]{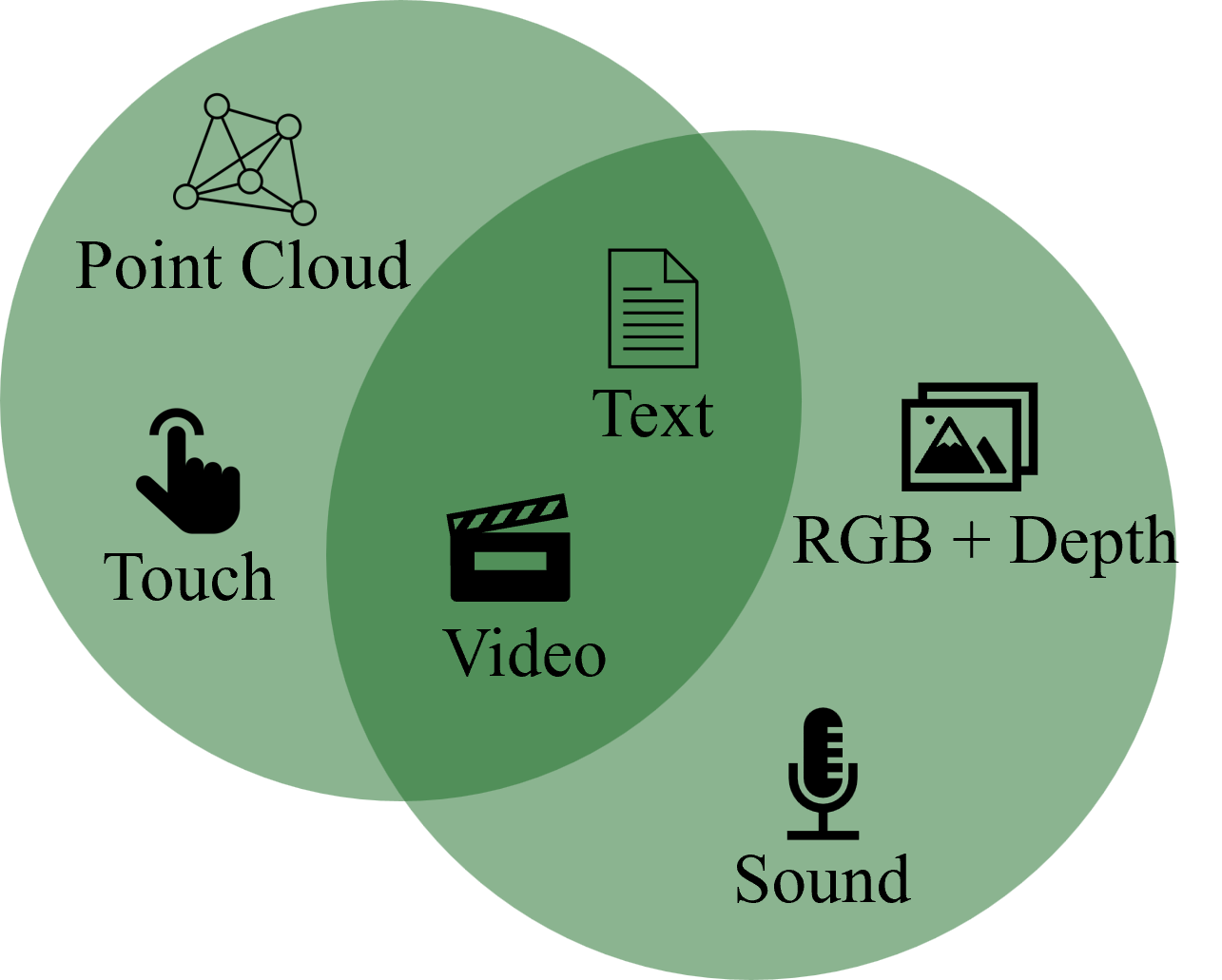}
        \caption{}
        \label{fig:how_to_1}
    \end{subfigure}
    \hfill
    \begin{subfigure}[b]{0.23\textwidth}
        \centering
        \includegraphics[width=\textwidth]{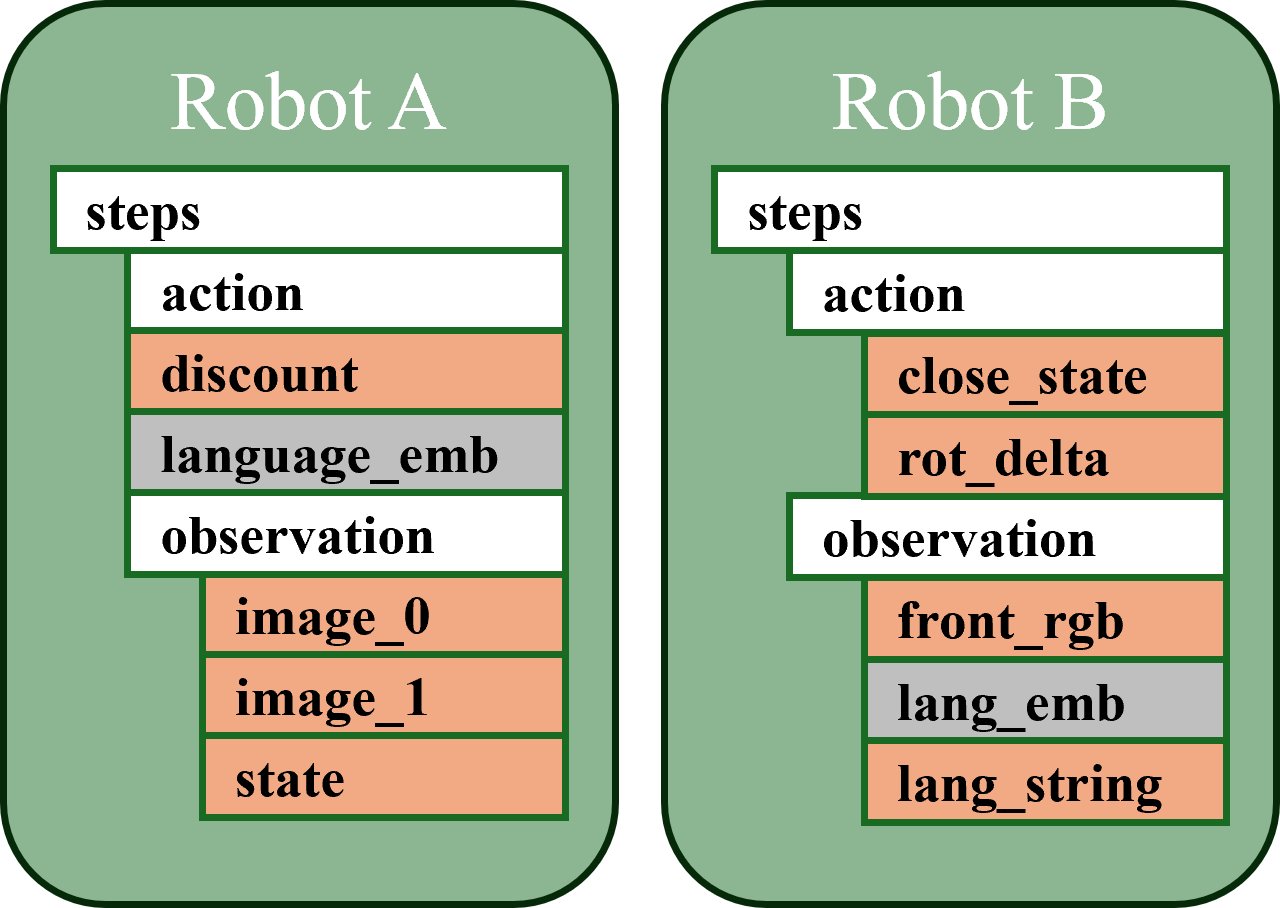}
        \caption{}
        \label{fig:how_to_2}
    \end{subfigure}

    \begin{subfigure}[b]{0.23\textwidth}
        \centering
        \includegraphics[width=\textwidth]{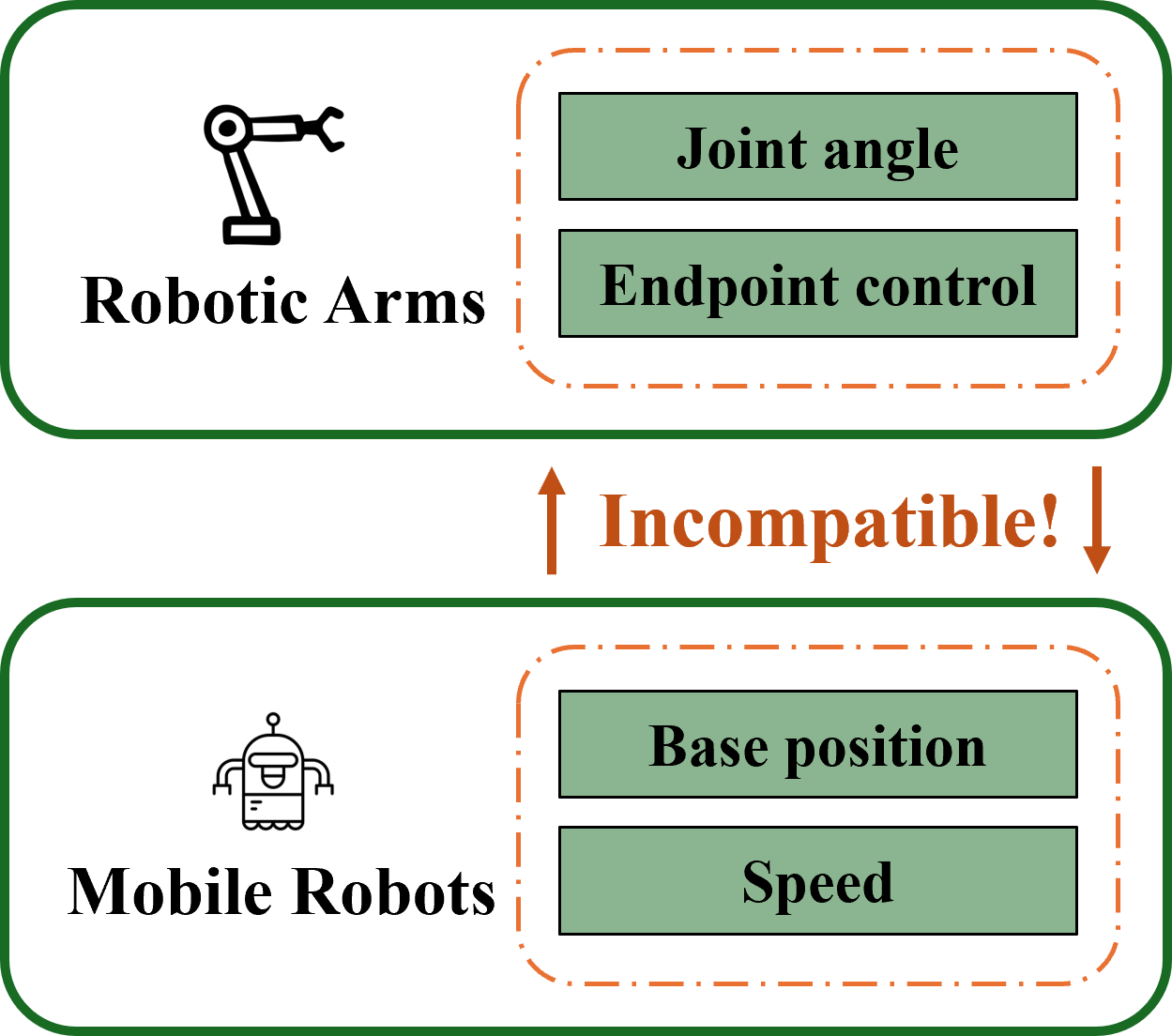}
        \caption{}
        \label{fig:how_to_3}
    \end{subfigure}
    \hfill
    \begin{subfigure}[b]{0.23\textwidth}
        \centering
        \includegraphics[width=\textwidth]{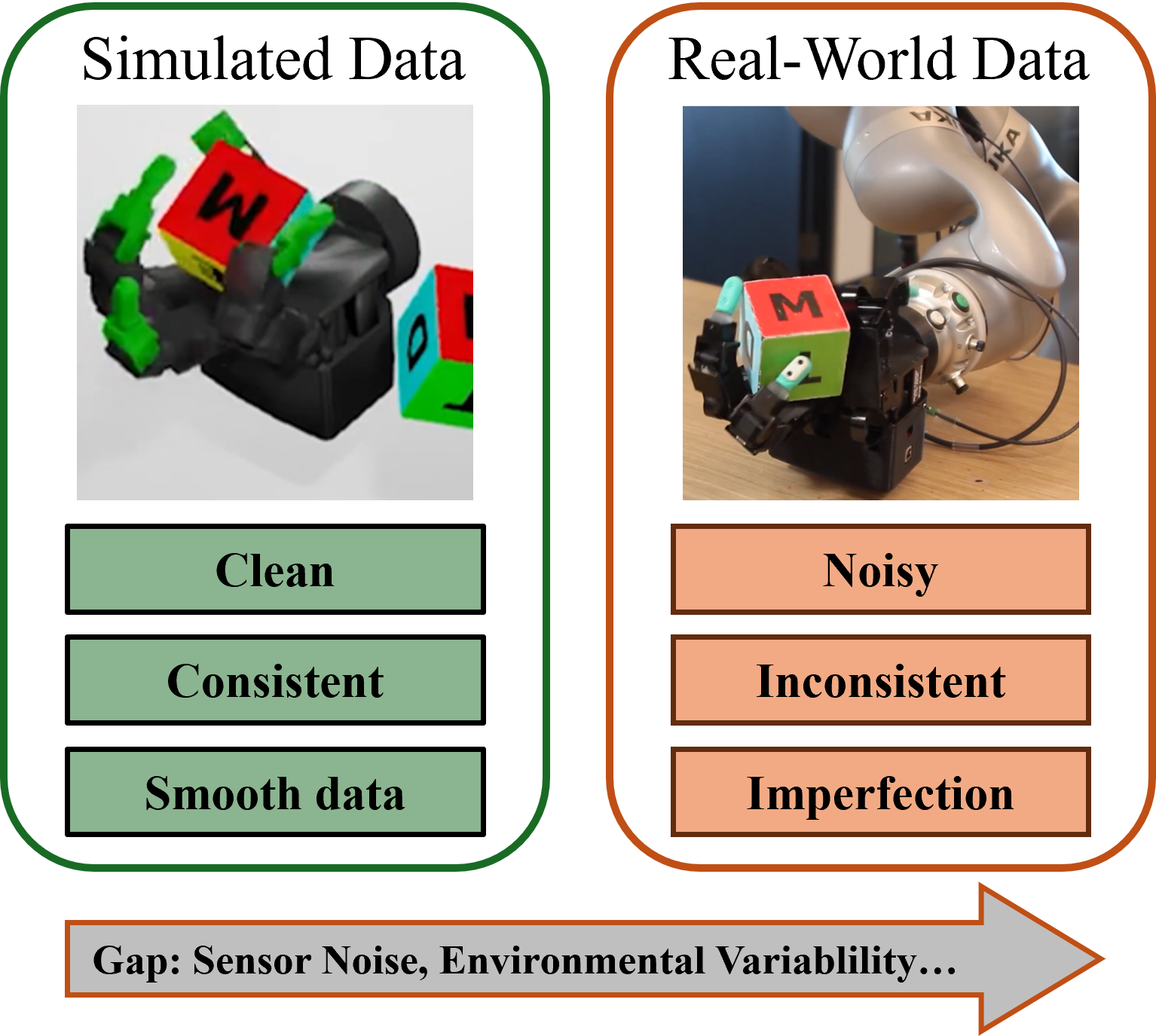}
        \caption{}
        \label{fig:how_to_4}
    \end{subfigure}
    
    \caption{Challenges in embodied intelligence datasets: (a) Insufficient sensory modalities, lacking comprehensive data across multiple input types; (b) Lack of standardization, complicating data processing across diverse robotic forms; (c) Incompatibility across platforms, hindering unified control of various robot types; (d) The gap between simulation and reality, highlighting the need for integrated datasets bridging simulated and real-world data.}
    \label{fig:challenges}
\end{figure}

\subsection{Why do we need ARIO?}

Embodied intelligence poses unique data challenges that are essential for developing large-scale models. Traditional data types like text and images can be easily harvested from the internet; however, embodied intelligence data necessitates real robots executing tasks in specifically designed environments or through sophisticated simulations, such as those enabled by MuJoCo or Isaac Sim. This method incurs substantial time, cost, and computational demands, with the aggregation of millions of data points representing a significant challenge. Collaborative initiatives, like the Open X Embodiment\cite{open_x_embodiment_rt_x_2023}, seek to consolidate datasets from various sources into a single platform to foster substantial and diverse data collections.

Embodied intelligence data is characterized by two primary features:
\begin{itemize}
    \item \textbf{Diverse Robotic Forms:} Robots exhibit a variety of configurations including single-arm, dual-arm, humanoid, and wheeled types. Each form demands unique control and motion data formats—ranging from joint angles to absolute positions. Unlike simpler data types, there is no universal data format that suits all these diverse forms.
    \item \textbf{Temporal Data Requirements:} Data frames must ideally be timestamped to properly sequence sensory inputs and control outputs. The variability in sensor frame rates and robot action frequencies adds another layer of complexity. When datasets from disparate sources are merged, each with its unique data storage conventions, it substantially complicates the processing and utilization of such data.
\end{itemize}

However, several critical shortcomings in existing datasets impede the progress in embodied intelligence:
\begin{enumerate}
    \item \textbf{Insufficient Sensory Modalities:} Current datasets lack richness in sensory modalities. No dataset currently encompasses a full array of data types including images, 3D structures, text, tactile, and auditory inputs simultaneously, as illustrated in Figure \ref{fig:challenges} (a).
    \item \textbf{Lack of Standardization:} While datasets like Open X Embodiment include diverse robotic forms, they suffer from a lack of a unified format, making it cumbersome to process and utilize, demonstrated in Figure \ref{fig:challenges} (b).
    \item \textbf{Incompatibility Across Platforms:} Existing datasets fail to accommodate diverse control objects in a unified format—ranging from navigation for mobile robots to the positional attitudes and joint angles for robotic arms, as shown in Figure \ref{fig:challenges} (c).
    \item \textbf{Gap Between Simulation and Reality:} There is a notable absence of datasets that offer both simulated and real data for the same robot, as well as simulated data based on real-world scans, which is crucial for studying the simulation-to-reality gap. This gap is highlighted in Figure \ref{fig:challenges} (d) and is a focus for ARIO, although currently unaddressed in available datasets.
\end{enumerate}

Due to these reasons, a dataset for embodied intelligence not only needs to be voluminous and varied but also standardized in a format that can adapt to the varying data variables of different robot morphologies with precise timestamping. This standardization facilitates the efficient training of high-performance, generalizable models for embodied applications. ARIO is designed to address these needs by offering an optimal data format standard for embodied intelligence. We have meticulously crafted the ARIO data structure to meet these criteria, incorporating both real and simulated data, and have made significant efforts to convert existing open-source datasets into the ARIO format, effectively tackling the current bottlenecks in the field of embodied intelligence.

\begin{figure*}[t]
  \centering
   \includegraphics[width=0.9\linewidth]{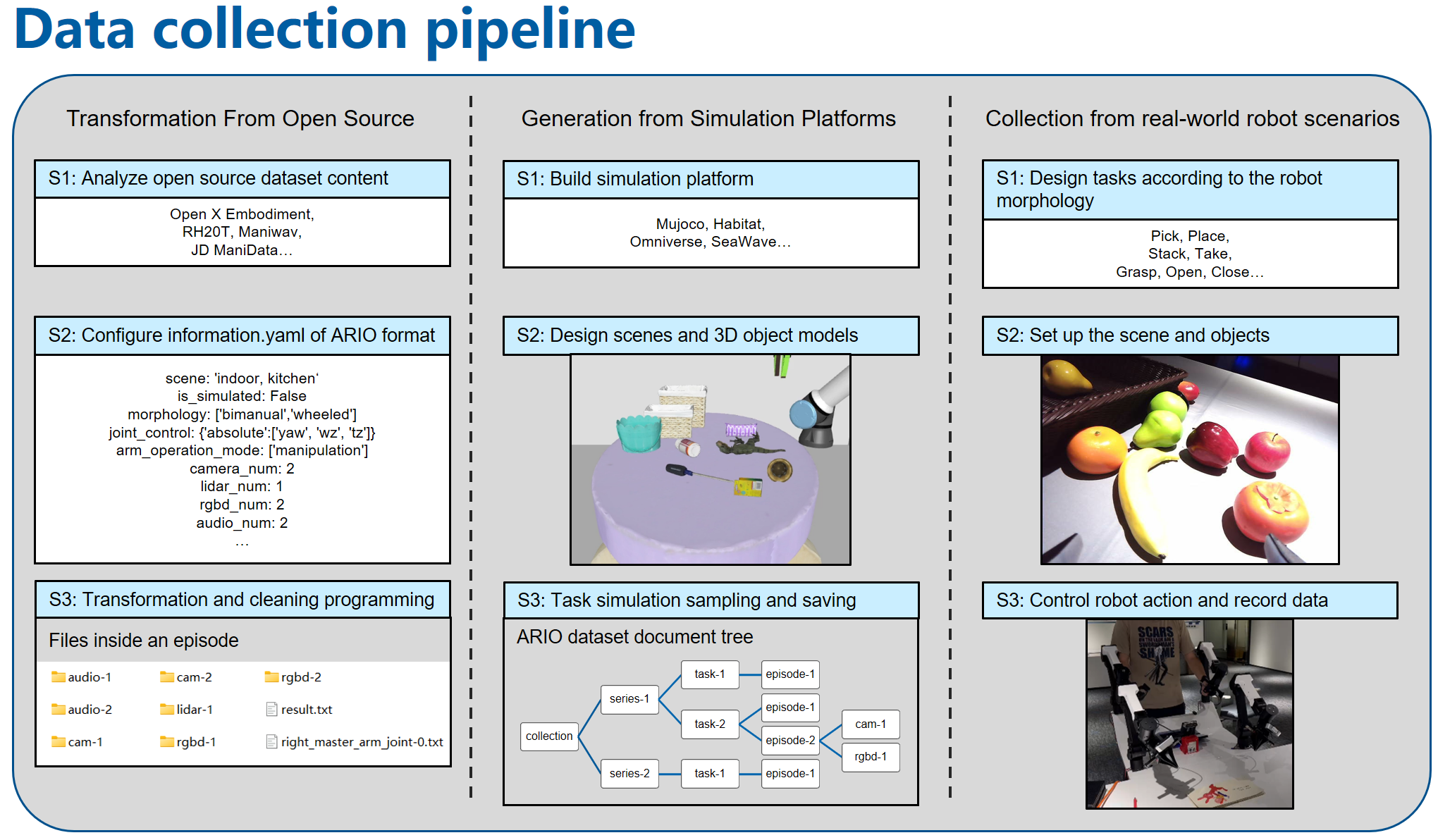}

   \caption{Collection pipeline of ARIO.}
   \label{fig:collection_pipeline}
\end{figure*}

\begin{table*}[htp]
\centering
	\fontsize{10}{10}\selectfont
	\caption{Comparison between our ARIO dataset and existing embodied datasets. 
 \vspace{-0.2cm}
	}
		\resizebox{1.0\linewidth}{!}{
	{\renewcommand{\arraystretch}{1.1}	\begin{tabular}{cccccccc}
				\specialrule{.2em}{1.5pt}{1.5pt}	
	Dataset&Modality&Task&Data Scale&Data Sources&Diverse Robotic Forms&Unified Format&Time-stamp Record
\\ 
 \specialrule{.1em}{0.8pt}{0.8pt}
         JD ManiData&2D, 3D, text&manipulation&0.5K&real&\XSolidBrush&\Checkmark&\XSolidBrush\\
         Maniwav&2D, text, audio&manipulation&1.3K&real&\XSolidBrush&\Checkmark&\XSolidBrush\\
          RH20T&2D, 3D, text, tactile&manipulation&13K&real&\Checkmark&\Checkmark&\Checkmark\\
          DROID&2D, 3D, text&manipulation&92K&real&\XSolidBrush&\Checkmark&\Checkmark\\
          Open-X Embodiment&2D, 3D, text&manipulation&2.4M&sim, open-source, real&\Checkmark&\XSolidBrush&\XSolidBrush\\
 \specialrule{.1em}{0.8pt}{0.8pt}
        ARIO&2D, 3D, text, tactile, audio&manipulation \& navigation&3M&sim, open-source, real&\Checkmark&\Checkmark&\Checkmark\\
         \specialrule{.2em}{0.8pt}{0.8pt}

\end{tabular}}}
\vspace{-0.2cm}	
\label{tab:statistic-total}
\end{table*}

\subsection{How do we build ARIO dataset?}
The development of ARIO aims to create a unified dataset for research on embodied agents, facilitating systematic progress in this field. The design requirements of ARIO are detailed in~\ref{subsubsec:design req}. The ARIO data collection pipeline consists of 3 paralleled components: collection from real-world scenarios (\ref{subsubsec:real-data}), generation from simulation platforms (\ref{subsubsec:sim-data}), and Transformation from open source datasets (\ref{subsubsec:transform-data}), as shown in Figure.~\ref{fig:collection_pipeline}.

\subsubsection{Design requirements}
\label{subsubsec:design req}
The issues discussed in the previous section lead to the following requirements that we seek to fulfill.
\begin{itemize}
  \item
Multiple sensory modalities. ARIO supports 5 modalities, including 2D image, 3D vision, sound, text, and tactile.

  \item
Temporal alignment of multi-modality data. ARIO supports timestamp-based recording and naming for camera (30Hz), lidar (10Hz), proprioception (200Hz) and tactile (100Hz) alignment.

  \item
Unified data architecture. ARIO is built on a scene-task-episode structure with textual descriptions of each scene and task, documenting rich information in an organized and logical manner.

  \item
Unified configuration. ARIO specifies the data content in the form of configuration files in a uniform format, enabling flexible recordings of data for many types of robot embodiments (single-arm, dual-arm, humanoid, quadruped, mobile) and different control actions (position, orientation, velocity, torque, \textit{etc.}).

 \item
Enhanced data diversity. ARIO thrives on a wealth of data resources: 1) It integrates data from both simulated environments and real-world scenarios; 2) It encompasses a broad spectrum of embodied tasks, featuring an array of robotic manipulation and navigation tasks; 3) It covers multiple robot morphologies and environments with high generalizability, which fascilitates research endeavors in cross-embodiment learning; 4) It preprocesses extensive open-source datasets, standardizing formats to ensure seamless integration and usability.
  
\end{itemize}
\subsubsection{Collection from real-world scenarios}

\label{subsubsec:real-data}

\begin{figure}[t]
  \centering
   \includegraphics[width=0.8\linewidth]{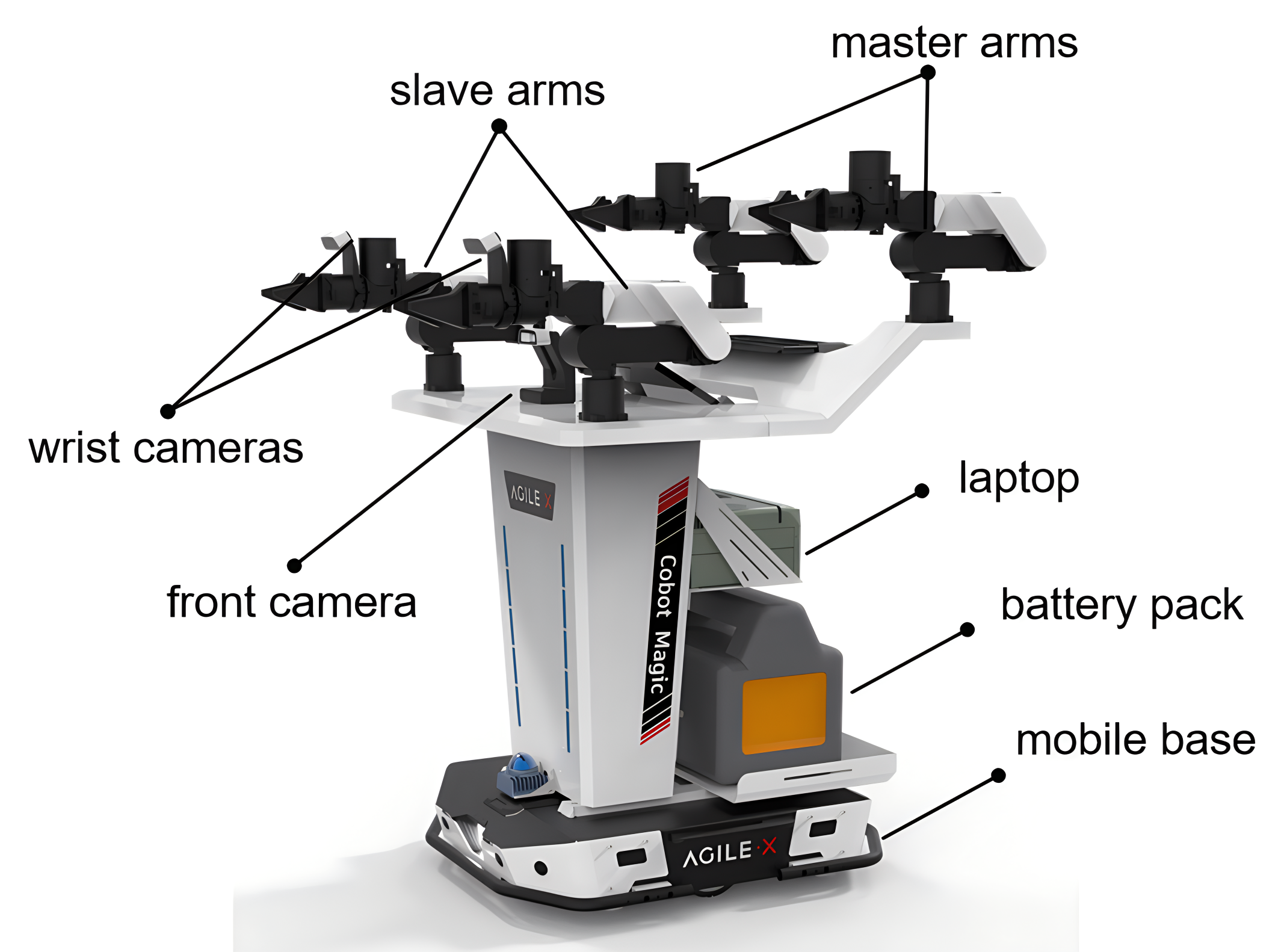}

   \caption{Illustration of the data collection platform which supports bimanual and whole-body teleoperation.}
   \label{fig:platform}
\end{figure}

\textbf{Cobot Magic data collection platform}.
We adopt \textit{Cobot Magic} (AgileX Robotics), a bimanual mobile manipulation plaftform ( Figure~\ref{fig:platform}) for our real-world data collection. It features a mobile base (AgileX Tracer AGV) for navigating around the environment with a speed up to 1.6 m/s. It also consists of 4 lightweight 6-dof arms (ARX ROBOTICS) with a tip payload of 3Kg, where 2 of them are master arms that can be controlled by human demonstrators intuitively through gravity compensation, the other two are slave arms that follow the movement of the master arms faithfully. The arms can be force-controlled and provide coarse haptic feedback by estimating the tip force through joint currents. A custimized gripper controlled by a linear motor is placed at the tip of each arm, with the two master arm grippers containing additional ``handle" mechanisms for continous control of the gripper by the operator. A total of 3 RGB-D cameras  (Orbbec DaBai) are adopted to provide ``observations" of the tasks, where two are mounted at the wrist of the slave arms, and the third one is placed at the front of the stand, facing forward.

The rest of the platform includes a battery pack and an onboard laptop for computation. The laptop accepts streaming from the 3 RGB-D cameras at resolution of 640$\times$480 and at 30Hz. It also accepts proprioception streaming from all 4 arms through
USB serial ports, and from the Tracer mobile base
through CAN bus. We record the joint positions, velocities and torques of all 4 robot arms at 200Hz. In addition, 6D pose of the end-effector and the state of the gripper are recorded at 200Hz. The data collection code is from \url{https://github.com/agilexrobotics/ario-tools}, which also supports recordings of point cloud and base velocities if needed. 

\textbf{Cobot Magic data collection and processing}. 
We recruit more than 30 volunteers to teleoperate the designed tasks. Volunteers receive instructions and detailed descriptions of the tasks that are about to collect, which also include the initial and final conditions of the robot and environment. They familiarize themselves with the task's operation before collecting data. For each task, 50 episodes are collected, where intial and final condition of the robot/environment may be varied  across different episodes. The volunteers are also required to classify the task's difficulty level as easy, medium, or hard after completing each manipulation.

The data collected with the collection software doesn't adhere to ARIOS's format strictlly, so additional postproccessing is carried out. Moreover, manual validation is performed to filter out episodes with missing information and reduced camera frequency.

\textbf{Cobot Magic data task design}. 
We design over 60 tasks, featuring table-top manipulation in household settings. The tasks cover not only general \textit{pick and place} skills, but also more complex skills like \textit{twist}, \textit{insert}, \textit{press}, \textit{cut}, etc.. In contrast to generic grasping tasks, we are particularly focused on the subsequent categories of tasks: 
\begin{itemize}
  \item
  \textbf{Long-horizon tasks}: tasks typically combine a sequence of simple short tasks. Examples inlcude: rearrange a list of items following the instruction, stack the cups one by one, pick up the blocks with both hands and then knock against each other. 

  \item
  \textbf{Bimanual and fine manipulation tasks}: tasks that require cooperation of both arms and usually contain fine manipulation. Examples include: twist open/close the cap of the bottle, pass an item from one hand to another, unzip the purse, grab the book with both hands for reading.

  \item
  \textbf{Contact-rich tasks}: tasks necessitate substantial physical engagement with the environment. Exmaples include: insert the plug into the socket, twist open/close the bottle, write letters on the paper with a pencil.

  \item
  \textbf{Human-robot collaboration tasks}: tasks requie humans and robots working together towards a common goal. Examples include: apply ointment to a person's skin, hand over an item to a person standby.

  \item
  \textbf{Deformable objects manipulation tasks}: tasks require manipulation of non-rigid objects where the force exerted on the objects has to be controlled. Exmaples include: fold towels along the diagonals and lay the tablecloth on the table.

\end{itemize}

\begin{figure*}[t]
  \centering
   \includegraphics[width=0.9\linewidth]{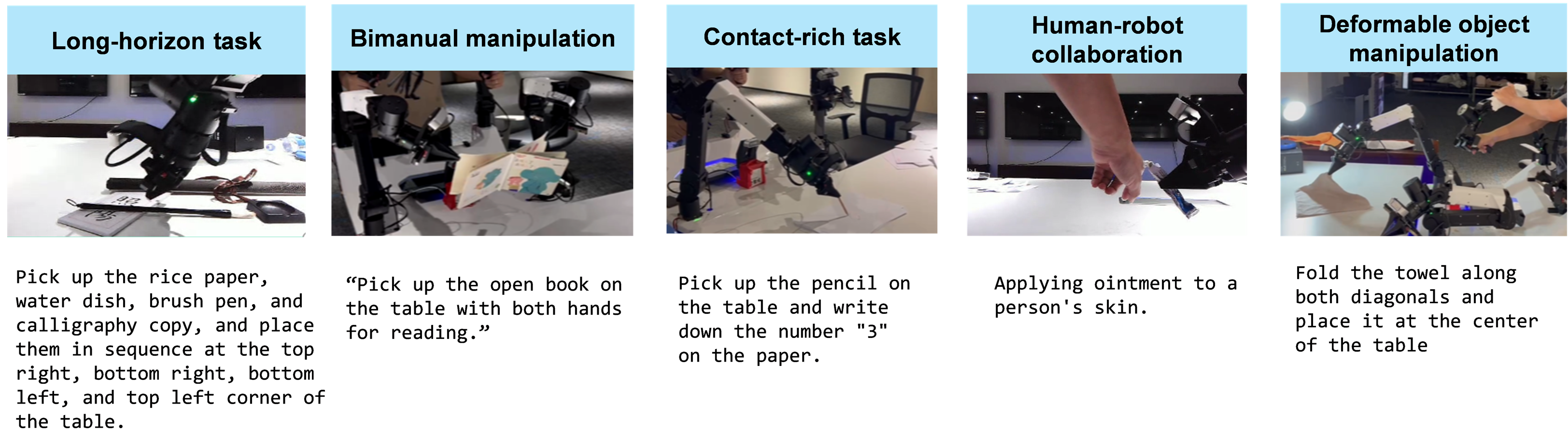}

   \caption{Some exmaple tasks, with the top row indicating the task category while the text at the botom row providing task instructions.}
   \label{fig:task}
\end{figure*}

We believe that human teleoperation can significantly enhance task execution and data quality for data collection of these tasks, comparing to scripted policies within simulated environments. It should be noted that some tasks may fall into multiple categories. Some example of the tasks can be viewed in Figure~\ref{fig:task}. For a complete list of tasks, please refer to our website:\url{https://imaei.github.io/project_pages/ario/}.

\textbf{Cloud Ginger XR-1 data collection platform}. We adopt Cloud Ginger XR-1 for our real-world data collection. Cloud Ginger XR-1 is a 5G-enabled wheeled humanoid cloud robot under XR series. Cloud Ginger has more than 40 smart joints with precise positioning, amorphous motor, hollow alignment design, high level of integration and large torque. Therefore, Cloud Ginger is highly flexible and can act in a more reliable manner, with higher load capacity. Cloud Ginger has multiple hardware expansion interfaces throughout the robot body, thus can incorporate multiple peripheral tools for a wider range of manipulation cases. In particular, Cloud Ginger has dexterous hands with seven degrees of freedom and a load of five kilograms when working together with arms to achieve dexterous grasping at high precision, which makes it easier to grasp, manipulate tools and perform other fine movements. In addition, Cloud Ginger has nine expandable interfaces over its body, which can be easily expanded to equip a variety of peripheral tools. 

\textbf{Cloud Ginger XR-1 data task design}. We designed three distinct tasks:
\begin{itemize}
    \item \textbf{Pick up}: The robot is required to pick up the correct object from the table.
    \item \textbf{Put on}: The robot picks up the object and places it on the corresponding color block.
    \item \textbf{Push to}: The robot pushes the object towards the designated color block.
\end{itemize}
For these tasks, we collected a total of 400, 200, and 200 episodes, respectively.

\textbf{Cloud Ginger XR-1 data collection and processing}. We employed human demonstration data collection, where operators directly control the robot to complete tasks. This approach allows us to capture expert-level demonstrations and ensure the collected data is aligned with the intended task objectives. Before formal data collection, an initial phase was conducted to familiarize ourselves with the operation. The preliminary data collected during this phase was utilized for training purposes to identify any potential issues and assess the utility of the data. Subsequently, the collected data underwent post-processing to convert it into the ARIO format.

\subsubsection{Generation from simulation platforms}
\label{subsubsec:sim-data}
 The simulation data in ARIO comes from three simulation platforms: object navigation tasks from Habitat, manipulation tasks from MuJoCo and SeaWave.

\textbf{Object Navigation (HM3D) task from Habitat}. We use Habitat-Matterport 3D (HM3D) scene datasets~\cite{ramakrishnan2hm3d} and Habitat Challenge 2022 Object Navigation (ObjectNav) task dataset for HM3D~\cite{habitatchallenge2022}. Habitat~\cite{savva2019habitat} is a platform for research in embodied AI, which enables training embodied agents (virtual robots) in highly efficient photorealistic 3D simulation. Habitat-Matterport 3D Research Dataset (HM3D)~\cite{ramakrishnan2hm3d} is the largest dataset of 3D indoor scenes. Object Navigation is a representative and challenging navigation task that focuses on egocentric object/scene recognition and commonsense understanding of object semantics. We collect navigation tasks of 6 goal categories following the setup in~\cite{habitatchallenge2022}, and each task is associated with a goal category. We use 80 scenes from the train split and the corresponding episodes to collect trajectories to find a given goal instance. We instantiate a shortest path greedy follower agent with official habitat API, which receives a given position and navigates to it through the shortest navigable path. For the input goal position, we use the position of the first goal instance in the scene goal objects list for most of the data and randomly chosen instances from the same list for the rest of the data. During each episode, the agent records the RGB-D observation and agent states at each timestep. We query the simulator for episodic success status and delete failed trajectories.

\textbf{Manipulation tasks from MuJoCo}. We adopt Scaling Up and Distilling Down~\cite{ha2023scalingup}, which develops a framework for Large Language Model (LLM)-guided task generation and policy learning, based on the MuJoCo physics simulator. Their framework utilizes UR5, a 6-degree-of-freedom robotic arm, to create an implementation API for a hierarchical control system. This system integrates high-level task policies, mid-level motion planning and low-level joint control. Based on this framework, we have designed three tasks: picking up objects, placing objects in baskets, and opening drawers. Corresponding scenes were constructed with 3D object models for each task. During the initialization phase of the sampler, we randomly positioned the target object and set its orientation within a predefined range. Additionally, several distractor objects were placed around the target to enhance task complexity. For the drawer-opening task, we also introduced variability by changing the background table color and the open/close state of unrelated drawers. These modifications increased scene diversity. We further recorded intermediate state data in ARIO format. In total, we generate 1,700 trajectory episodes involving 21 interactive objects.

\textbf{Manipulation tasks from SeaWave}. We convert the original SeaWave~\cite{ren2024surferprogressivereasoningworld} data into ARIO format. The SeaWave benchmark includes a simulator based on UE5 designed to evaluate robots' ability to understand and execute human natural language instructions. The simulator tests various robotic manipulation skills, such as pick, place, and move near, in different task scenarios. SeaWave tasks are classified into four levels of complexity based on the nature of the instructions and the required operations. These range from basic manipulation tasks with simple commands to complex scenarios requiring visual perception and decision-making based on abstract natural language instructions.

\subsubsection{Transformation from open source datasets}
\label{subsubsec:transform-data}

The transformation data in ARIO comes from three open-source datasets: Open X-Embodiment, RH20T, and Maniwav Datasets.

\textbf{Transformation based on the Open X-Embodiment dataset}.
The Open X-Embodiment dataset~\cite{open_x_embodiment_rt_x_2023} is a large-scale open-source resource that aggregates data from various institutions. It comprises 72 datasets, totaling over 2.4 million episodes, and showcases a diverse array of robots, control methods, and data collection strategies. The dataset includes a wide variety of tasks, such as grasping, collecting, classifying, and navigating, incorporating both real-world and simulated data.

However, the dataset has some notable drawbacks. It lacks a unified and clear format, making it challenging to process and load the data efficiently. To address this issue, we developed a conversion tool to transform the complex and diverse data in Open X-Embodiment into our proposed ARIO format. During the conversion process, we encountered missing data and unclear documentation, which made it difficult to ascertain the correct interpretation of some data. Consequently, we aimed to preserve the original data as much as possible while removing irrelevant or unusable data.

We filtered out datasets without gripper and joint information or those with unclear documentation, resulting in the conversion of 62 datasets.

\textbf{Transformation based on the RH20T dataset}.
The RH20T~\cite{fang2023rh20t} dataset is compiled from real-world teleoperated tasks in a variety of settings. The acquisition environment is equipped with an ensemble of global RGBD cameras ranging from eight to ten units, providing diverse viewpoints, along with one to two in-hand cameras. Comprehensive kinematic data, including joint angles and torques, as well as the gripper's position, orientation, and opening/closing status, are recorded. Certain robotic grippers are equipped with fingertip tactile sensors, enhancing the richness of the modality in contact-rich tasks.

 The RH20T dataset encompasses over 140 tasks, featuring common activities encountered in daily life. Each task has been executed by multiple operators using seven different robotic arms across various conditions, ensuring a high level of diversity. Considerable efforts were dedicated to the calibration of all sensors, aiming to ensure data consistency and reliability.

To utilize this dataset effectively, we developed a conversion program to translate the RH20T dataset into the ARIO format. However, the dataset exhibits limitations, primarily due to missing data in some episodes, including the absence of camera feeds or joint data. Furthermore, the documentation provided is insufficiently detailed, posing challenges in fully interpreting the dataset's implications.

In our conversion process, we strived to preserve the integrity of the original data, excluding only those episodes where the task description was entirely absent. Through this approach, we successfully converted a total of 12,719 episode trajectories.

\textbf{Transformation based on the Maniwav dataset}.
Currently, the majority of open-source embodied intelligence datasets predominantly encompass visual perception data, with a minority including tactile information, and virtually none recording auditory data during the agent's interaction with the environment. To the best of our knowledge, the ManiWAV~\cite{liu2024maniwav} team stands uniquely in investigating the influence of sound on robotic task success rates, having released a dataset that captures this crucial component.

The ManiWAV dataset features four tasks: Wiping Whiteboard, Flipping Bagel, Pouring Dice, and Taping Wires with Velcro Tape. Data collection was conducted using a human-operated Universal Manipulation Interface (UMI)~\cite{chi2024universal} device, which was equipped with a co-moving camera and dual microphones-one designated for capturing contact audio and the other for environmental sounds. Complementing these auditory recordings, the dataset also includes positional and orientational Euler angle data of the UMI gripper, alongside its opening and closing states.

We have developed a conversion script to transform all publicly available ManiWAV data into the ARIO format, augmenting each task with textual descriptions for enhanced clarity and utility. Encompassing a total of 1,297 episode trajectories, ManiWAV represents the sole dataset within ARIO that incorporates auditory data, offering a unique resource for research into multimodal robotic perception and action.

\subsubsection{Data statistics}
\label{subsubsec:data statistics}
The ARIO dataset collects a total of 258 series, 321,064 tasks, and 3,033,188 episodes, as shown in Table~\ref{tab:statistic-total}. In the following, we provide comprehensive statistics on all aspects of ARIO, including scenes, skills, data sources, and robot-related parameters. 

\begin{table}[ht]
\centering
	\fontsize{5}{5}\selectfont
	\caption{Statistics of the ARIO dataset. 
 \vspace{-0.2cm}
	}
		\resizebox{1.0\linewidth}{!}{
	{\renewcommand{\arraystretch}{1.1}	\begin{tabular}{cccc}
				\specialrule{.1em}{1pt}{1pt}	
	Data source&Series&Tasks&Episodes\\ 
 \specialrule{.1em}{0.8pt}{0.8pt}
        Transformation from open-source&161&319761&2326438\\
        Collection from real-world  (ours)&2&105&3662\\
        Generation from simulator (ours)&95&1198&703088\\ 
        \specialrule{.1em}{0.8pt}{1pt}
        Total&258&321064&3033188\\ 
         \specialrule{.1em}{0.8pt}{0.8pt}

\end{tabular}}}
\vspace{-0.2cm}	
\label{tab:statistic-total}
\end{table}

\input{Figures/scene_skill_distribution}

\textbf{Statistics on scenes}. Figure~\ref{fig:scene_skill_distribution} (a) shows that the ARIO dataset predominantly contains series in \textit{table} and multiple indoor scenes (\textit{open domains}, \textit{multi-room}, \textit{living room}, and \textit{kitchen}), with 165 and 213 series, respectively. From Figure~\ref{fig:scene_skill_distribution} (a), we can see that the scenes in ARIO with higher number of tasks are primarily in indoor environments such as \textit{tables}, \textit{kitchens}, \textit{households}, \textit{hallways}, and \textit{multi-room} settings. Figure~\ref{fig:scene_skill_distribution} (a) also shows the specific number of episodes under each scene type, which reaches a number in the millions in the \textit{table} scene. The ARIO dataset collects a variety of scene types, rich task instructions, and large-scale data to provide powerful data support for the development of general-purpose embodied agents.

\textbf{Statistics on skills}. Figure~\ref{fig:scene_skill_distribution} (b) shows the number of series, tasks and episodes corresponding to each skill in the ARIO dataset. The ARIO dataset encompasses a broad range of skills necessary for embodied manipulation and navigation and is equipped with a comprehensive set of task instructions and episodes. Each skill in the ARIO dataset is collected in diverse scenarios, facilitating the acquisition of scenario generalizability for powerful embodied agents.

\begin{figure*}[htp]
\begin{centering}
\includegraphics[width=1.0\linewidth]{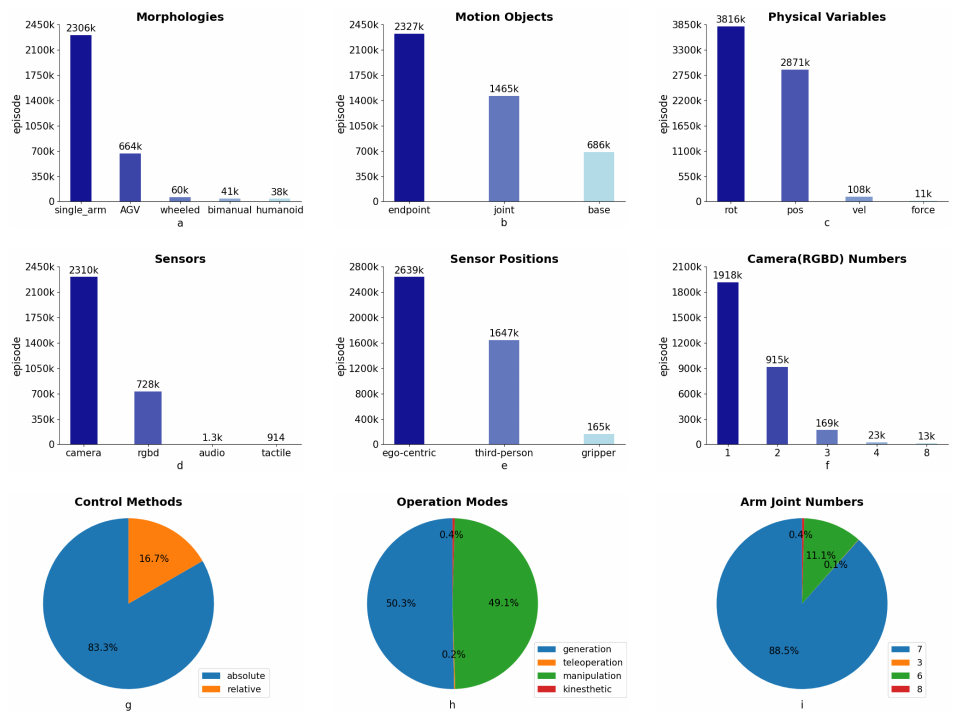}
\par\end{centering}
\caption{Statistics on robot parameters in the ARIO dataset.}
\label{fig:robot-param}
\vspace{-0.2cm}
\end{figure*}

\begin{figure}[h]
\begin{centering}
\includegraphics[width=1.0\linewidth]{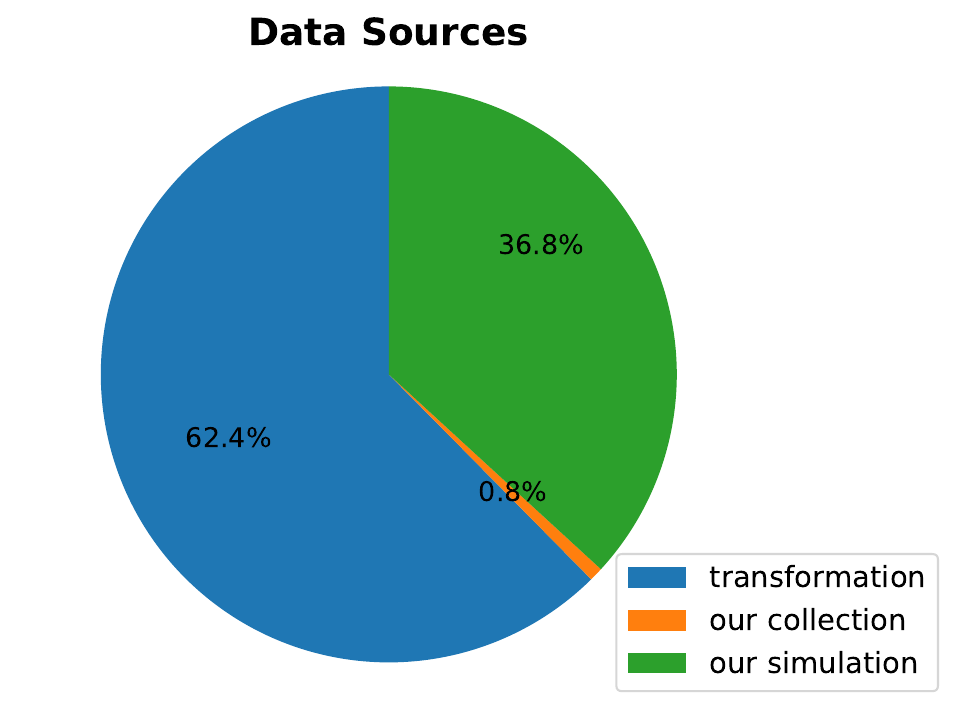}
\par\end{centering}
\caption{The proportion of data sources in the ARIO dataset.}
\label{fig:data-source-pie-series}
\vspace{-0.2cm}
\end{figure}

\textbf{Statistics on data sources}. The ratio of simulation, transformation, and real-world data in the ARIO dataset is shown in Figure~\ref{fig:data-source-pie-series}, and corresponding specific data statistics are in Table~\ref{tab:statistic-total}. Note that the data transformed from open-source datasets contains both simulation and real-world data, but the real-world and simulation data in Figure~\ref{fig:data-source-pie-series} and Table~\ref{tab:statistic-total} specifically refer to the data we collected. The diverse data sources in the ARIO dataset are beneficial for training generalized embodied agents.

\textbf{Statistics on robot parameters}. We present the episodic statistical results of the various robot-related parameters in the ARIO dataset. Figure~\ref{fig:robot-param} (a) counts the number of robot morphologies, corresponding to single-arm, AGV (navigation robot), wheeled, bimanual, and humanoid robots. Figure~\ref{fig:robot-param} (b) shows the number of robot motion objects, corresponding to endpoint, joint, and base. Figure~\ref{fig:robot-param} (c) counts the number of robot physical variables included in the dataset, rot denotes the number of episodes containing rotation angles, including joint angles and rotation angles of the arm ends or the body, pos denotes the number of episodes containing positions, including arm end positions or body positions, vel denotes the number of episodes containing velocities, including joint velocities and arm end or body's movement speed, force denotes the number of episodes containing force or moment, including joint's and end's. Figure~\ref{fig:robot-param} (d) shows the number of different sensors, corresponding to the camera, rgbd camera, audio sensor, and tactile sensor. Figure~\ref{fig:robot-param} (e) counts the number of different sensor positions in the dataset, ego-centric denotes a first-person perspective, which generally means that the camera or rgbd is mounted at the head position, third-person generally means that the sensor is mounted in the surroundings, and gripper, which denotes end-effector mounted, e.g., wrist camera, fingertip haptics. Figure~\ref{fig:robot-param} (f) counts the number of cameras. Figure~\ref{fig:robot-param} (g) gives the statistical robot control mode ratio, relative means that the robot's motion control instruction is based on the relative value of the current position, absolute means that it is based on the absolute value of a certain coordinate system, for example, the instruction x=0.1, relative means that it is moved 0.1 from the current position to the direction of x, and absolute means that it is moved to the position of x=0.1. Figure~\ref{fig:robot-param} (h) shows the distribution of robot operation modes in the dataset. Kinesthetic means that the human directly moves the robot to a specified position or pushes the robot to a specified position, such as Aloha's master hand, manipulation means that the human operates it through a remote control, teleoperation means that there are motion sensors on the human's hand or body to collect human movement data, and the robot follow the collected data, such as Aloha's slave hand, or controlled by VR equipment, imitation means that the robot learns/imitates human movements through sensors such as LIDAR/camera, and there is no direct contact between the robot and human, generation means that the robot's movement data is generated directly through algorithms or tools or simulation, and does not require human operation. Figure~\ref{fig:robot-param} (i) counts the proportion of the number of arm joints for robots with arms. Figure x illustrates that the ARIO standard  constitutes a unified dataset for various robot settings to support efficient generalized agent training.

%% file: Figures/scene_skill_distribution.tex
\begin{figure*}[htp]
  \centering
   \includegraphics[width=1.0\linewidth]{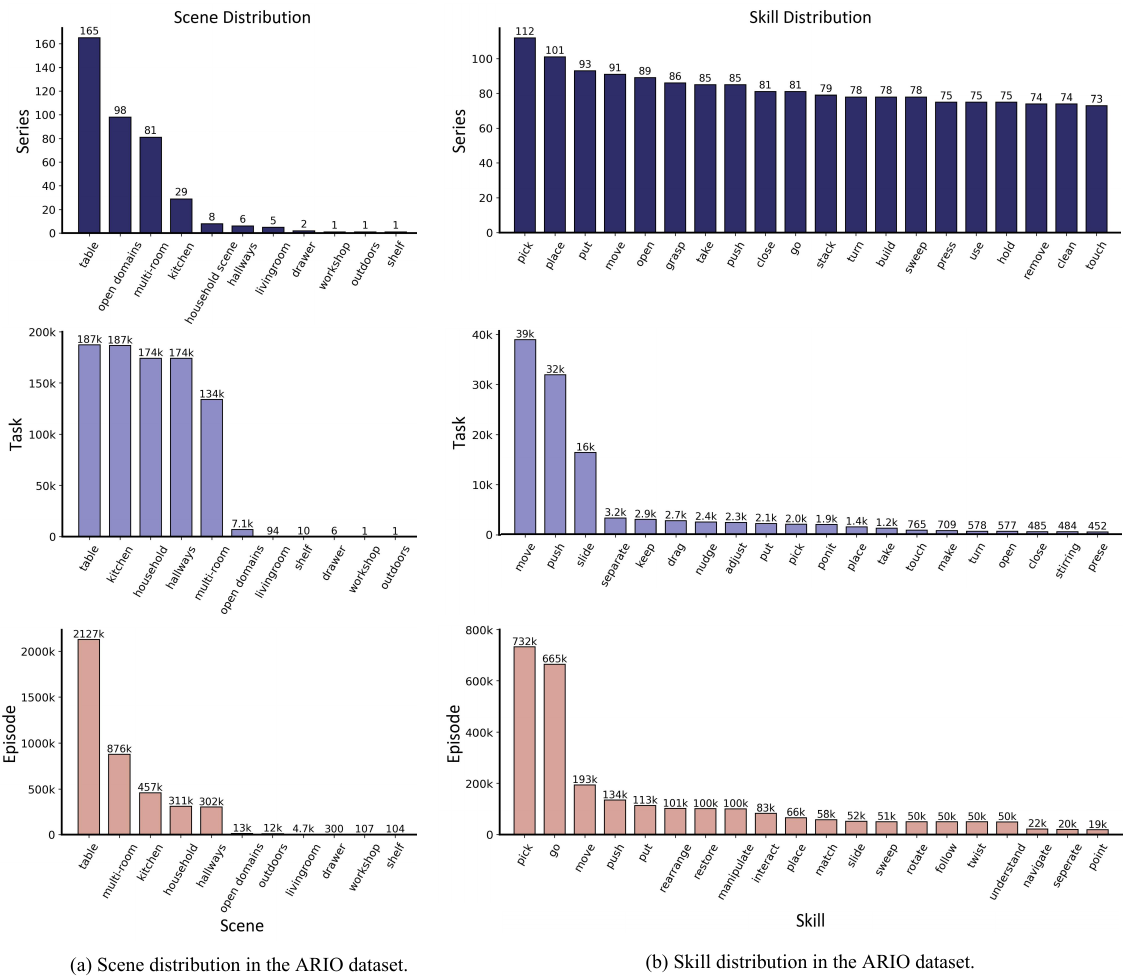}
   \caption{Statistics on scenes and skills in the ARIO dataset. (a) scene distribution. (b) skill distribution.}
   \label{fig:scene_skill_distribution}
\end{figure*}

%% file: S4_Discussion.tex
\section{Future work}
This work lays the foundation for future research in embodied AI, with several promising avenues for exploration:

\begin{itemize}
    \item \textbf{Evaluating ARIO for Large-Scale Model Training.} A key next step is to investigate the effectiveness of ARIO for training large-scale embodied AI models. 
    This can be achieved by leveraging existing open-source models or developing specialized architectures, and evaluating their performance on a variety of tasks using the ARIO dataset. This evaluation will shed light on the potential of ARIO to enhance the generalization and robustness of embodied AI agents.

    \item \textbf{Expanding Data Diversity.} Further enriching the ARIO dataset is crucial for advancing embodied AI research. 
    This includes:
    (1) Multi-modal data collection: expanding the dataset to include richer sensory modalities, such as tactile data, audio signals, and point clouds. 
    This will enable the development of agents capable of perceiving and interacting with the world in a more comprehensive and nuanced manner.
    (2) Real-world and simulation alignment: focusing on collecting data from the same robots in both simulated and real-world settings. 
    This will facilitate the development of techniques for bridging the simulation-to-reality gap, leading to more reliable and transferable embodied AI agents.

    \item \textbf{Scaling Up Embodied AI Data.} To truly unlock the potential of embodied AI, we need to generate massive datasets that reflect the complexity of real-world scenarios. 
    This can be achieved by:
    (1) Constructing large-scale simulated environments: developing virtual environments populated with diverse robots, each performing different tasks. These environments can be used to generate millions or even billions of data points, providing a scalable and cost-effective way to train embodied AI models.
    (2) Multi-robot interaction: focusing on scenarios involving multiple robots interacting with each other and the environment. 
    This will allow for the development of AI agents capable of collaborating and coordinating their actions, paving the way for more complex and sophisticated embodied AI systems.

\end{itemize}

By pursuing these research directions, we aim to advance the field of embodied AI, enabling the development of more capable, adaptable, and intelligent agents that can seamlessly interact with the physical world.

%% file: S5_Acknowledgement.tex
\section*{Acknowledgement}
We extend our gratitude to the various open-source datasets and platforms, including Open X-Embodiment, RH20T, ManiWAV, JD ManiData, and the contributors from Open X-Embodiment. Their contributions were vital in creating the ARIO dataset.
Special thanks to the Habita-sim simulation platform, Habitat-lab module library, Habitat-Matterport 3D Dataset (HM3D) indoor dataset, and the Habitat Challenge organized by Facebook AI Research. It is through your open-source support that we were able to collect navigation simulation data.
We are grateful for the Scaling Up and Distilling Down project for the simulation framework and the MuJoCo physics engine, aiding in generating simulation manipulation data.
We appreciate the ARIO Embodied Intelligence Data Open Alliance members, such as Southern University of Science and Technology, Sun Yat-sen University, Dataa Robotics, Agilex Robotics, and JD Technology, for their technical support and contributions to the ARIO dataset development.
The collaborative efforts have significantly advanced embodied AI research through the creation of the ARIO dataset.